# A derivative-fidelity failure mode in physics-informed neural networks: strengthened benchmark evidence from function-value training

[Koji KOYAMADA][1],*

1 Osaka-seikei University

* [koyamada@g.osaka-seikei.ac.jp]

**Abstract.** Physics-informed neural networks (PINNs) use automatic differentiation to impose differential-equation residuals, but good agreement in function values does not necessarily imply accurate derivatives. This paper formulates derivative fidelity as a failure mode of PINNs and tests it with one-dimensional benchmarks. Multilayer perceptrons are trained only on function values for sin(x) and exp(x), while second derivatives obtained by automatic differentiation are evaluated separately. The hypothesis is strengthened by additional tests over training-point density, activation functions, endpoint-dense evaluation, and both L2 and maximum-error diagnostics. The results show that visually accurate function approximation can coexist with substantially larger second-derivative errors, especially near high-curvature boundary regions. The experiment provides a diagnostic protocol for distinguishing value accuracy from physics-residual reliability.

**Keywords:** Physics-informed neural networks, derivative fidelity, failure mode, automatic differentiation, function approximation, simulation reliability

## 1. Introduction

Physics-informed neural networks (PINNs) have become a widely used framework for solving forward and inverse problems governed by differential equations [1]. Their appeal is that a neural network surrogate can be trained not only from observations but also from residuals of governing equations evaluated through automatic differentiation. This construction is attractive for simulation because it appears to combine flexible function approximation with physical constraints.

However, a growing body of work has reported PINN failure modes: slow convergence, ill-conditioned loss landscapes, unbalanced gradients among loss terms, and solutions that match boundary or data values while violating the intended physics [2, 3]. These failures are often discussed as optimization or loss-balancing problems. This paper focuses on a complementary and more elementary diagnostic issue: a neural network may approximate target function values well while its differentiated output is a less accurate approximation of the target derivative.

Recent work has suggested that low PDE residuals or visually plausible solution fields do not necessarily guarantee preservation of the underlying physical structure. In particular, a recent comparison between standard PINNs and Trefftz-based PINNs [9] demonstrated that solutions with similarly small residual losses may exhibit substantially different streamline and magnetic-field topologies, indicating that physically important structures can be distorted even when conventional error metrics appear satisfactory [9].

This observation raises a more fundamental question. One possible explanation is that accurate approximation of function values does not necessarily imply accurate approximation of derivatives. If derivative fidelity is insufficient, the residual itself may become unreliable, potentially leading to erroneous physical structures. Motivated by this question, the present work investigates derivative fidelity in its simplest form using one-dimensional benchmark functions trained only on function values, thereby isolating the relationship between function-value accuracy and derivative accuracy before considering full PDE problems.

The distinction is important because many PINN residuals depend on first or second derivatives rather than on function values alone. Therefore, a visually convincing approximation of u(x) may still lead to a poor approximation of Lu(x), where L is a differential operator. This paper examines this issue with small benchmark functions before considering full PDE solvers. Accordingly, this work should be viewed as a foundational study aimed at understanding a possible mechanism behind previously observed structure-preservation failures.

# 2. Research question and hypothesis

The research question is: when a neural network is trained to approximate only the values of a smooth target function, does the automatically differentiated second derivative reliably approximate the analytical second derivative?

The working hypothesis is negative. Classical approximation results establish broad expressive capacity for neural networks [4, 5], but universal approximation at the level of function values, and a low value-based training loss in a finite experiment, do not by themselves guarantee derivative fidelity. In particular, high-order derivative errors may remain large in localized regions even when function-value plots appear accurate. For PINNs, this gap can appear as a hidden failure mode because the PDE residual is evaluated through derivatives.

# 3. Method

## 3.1. Benchmark functions

Two one-dimensional smooth targets were selected: f(x) = sin(x) on $[-2\pi, 2\pi]$ and f(x) = exp(x) on [-2, 2]. Their second derivatives are f''(x) = -sin(x) and f''(x) = exp(x), respectively. The sinusoidal case tests periodic curvature, while the exponential case tests monotone growth and rapidly increasing curvature near the right boundary.

## 3.2. Neural approximation and derivative evaluation

A fully connected multilayer perceptron was trained with a value-only mean-squared-error loss,

$$L_{value}(\theta) = \frac{1}{N}\sum_{i} |u_\theta(x_i) - f(x_i)|^2 \quad (1)$$

After training, the value error $e_0(x) = |u_\theta(x) - f(x)|$ and the second-derivative error $e_2(x) = |d^2u_\theta/dx^2 - f''(x)|$ were evaluated on dense grids. The second derivative of the network was not fitted directly in the primary experiments. It was obtained by automatic differentiation, matching the mechanism used in PINN residual evaluation.

## 3.3. Strengthened validation experiments

To strengthen the hypothesis test beyond a single training run, four additional comparisons were conducted: (i) a training-point sweep with $n_{train}$= 32, 64, 128, 256, and 512; (ii) an activation comparison among tanh, sine, and gelu; (iii) an endpoint-dense evaluation grid for exp(x), with region-wise error measurement; and (iv) a unified comparison using both RMSE and maximum absolute error. These tests were designed to distinguish average approximation quality from localized derivative failure.

Table 1: Experimental configuration

| Item | Setting |
|---|---|
| Architecture | Fully connected MLP, 3 hidden layers, width 64 |
| Activations | tanh for baseline; tanh, sine, and gelu for activation comparison |
| Training points | Baseline 128; sweep 32, 64, 128, 256, 512 |

| Evaluation | Uniform dense grids; endpoint-dense grid for exp(x) near x = 2 |
|---|---|
| Optimizer | Adam |
| Epochs | 2500 in strengthened validation experiments |
| Loss | Function-value MSE only |
| Metrics | Function RMSE/max error and second-derivative RMSE/max error |

# 4. Results

## 4.1. Baseline case: sin(x)

For sin(x), the learned function closely overlaps the exact curve across the full domain. The absolute function-value error is mostly in the range of approximately $1 \times 10^{-3}$ or below according to the plotted semilog profile. In contrast, the second-derivative error is visibly larger, reaching approximately $1 \times 10^{-2}$ to $1 \times 10^{-1}$ in localized regions. Thus, value accuracy and curvature accuracy are not equivalent even for a smooth periodic target.

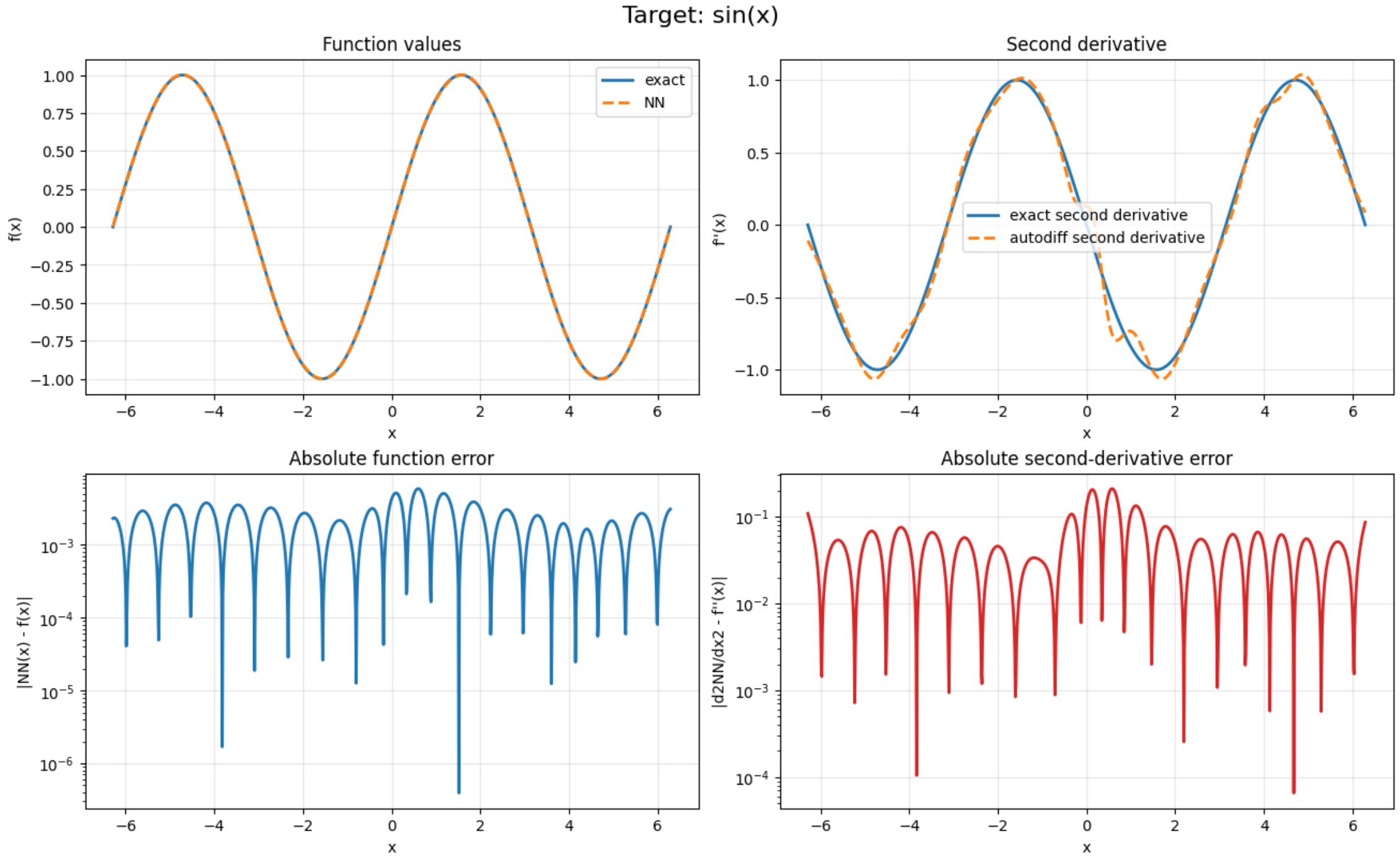

Fig. 1: Function approximation and second-derivative error for sin(x).

## 4.2. Baseline case: exp(x)

For $\exp(x)$, the function-value approximation is again visually accurate. The second derivative, however, deteriorates near the right boundary. Around x = 1.6 and beyond, the auto-differentiated second derivative deviates from the exact $\exp(x)$ curve, and near x = 2 the derivative error increases by orders of magnitude relative to most of the interior. This is the clearest initial instance of the proposed failure mode.

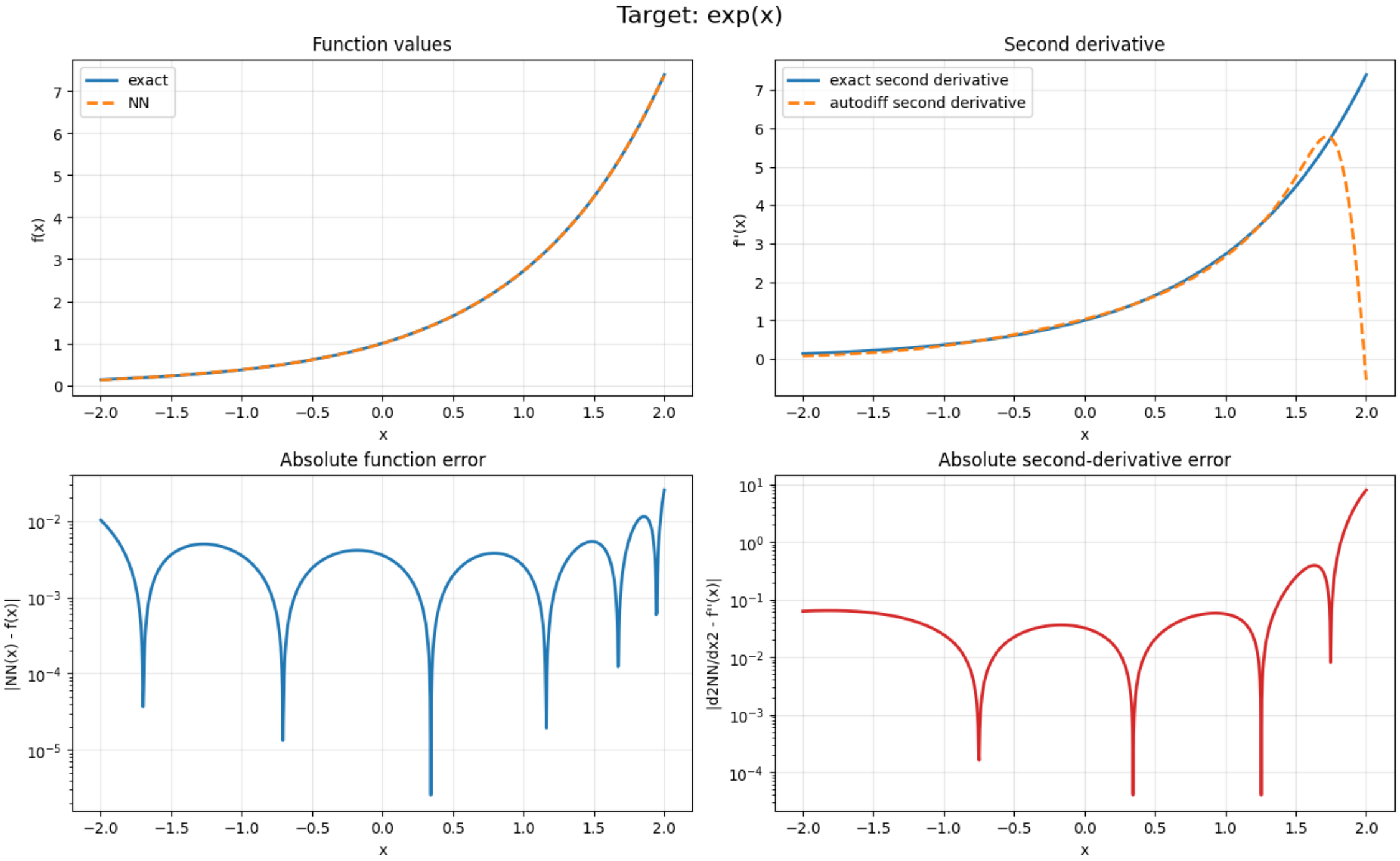

Fig. 2: Function approximation and second-derivative error for exp(x).

Table 2: Summary of baseline observed behavior

| **Target** | **Function-value fit** | **Second-derivative behavior** | **Interpretation** |
|---|---|---|---|
| sin(x) | The exact and learned curves nearly overlap. | The curvature is broadly reproduced but localized errors are larger than value errors. | Derivative fidelity is weaker than value fidelity. |
| exp(x) | The learned curve remains visually close to the exact function. | The second derivative collapses near the high-curvature right boundary. | Boundary and curvature-sensitive derivative failure is observed. |

## 4.3. Extended validation over data density, activation, and locality

The additional validation experiments strengthen the failure-mode interpretation in four ways. First, increasing $n_{train}$ did not monotonically reduce second-derivative error for $\exp(x)$: the second-derivative RMSE remained on the order of 1 and the maximum absolute error remained near 1e1 across the sweep. Second, activation choice substantially affected derivative fidelity. For sin(x), sine activation produced the lowest second-derivative RMSE, while tanh produced the largest derivative distortion. For exp(x), sine also reduced the maximum second-derivative error relative to tanh, but a large endpoint error remained. Third, endpoint-dense evaluation showed that the exp(x) collapse is localized primarily at the right high-curvature boundary rather than being a generic boundary artifact. Fourth, maximum absolute error revealed localized failures that are partly obscured by average metrics.

Table 3: Key strengthened-validation metrics

| **Comparison** | **Condition** | **Function RMSE** | **Second-derivative RMSE** | **Second-derivative max error** |
|---|---|---|---|---|
| $n_{train}$, sweep, exp(x) | best d2 RMSE at n=128 | 0.005033 | 0.847285 | 6.909722 |
| $n_{train}$ sweep, exp(x) | largest d2 max at n=256 | 0.003435 | 1.266890 | 10.714455 |
| activation, sin(x) | sine | 0.002062 | 0.022970 | 0.116833 |
| activation, exp(x) | sine | 0.004452 | 0.578200 | 3.931784 |
| activation, exp(x) | tanh | 0.012775 | 0.831892 | 6.212383 |
| endpoint dense, exp(x) | right boundary / high curvature | (small visually) | dominant region | approximately 10 |

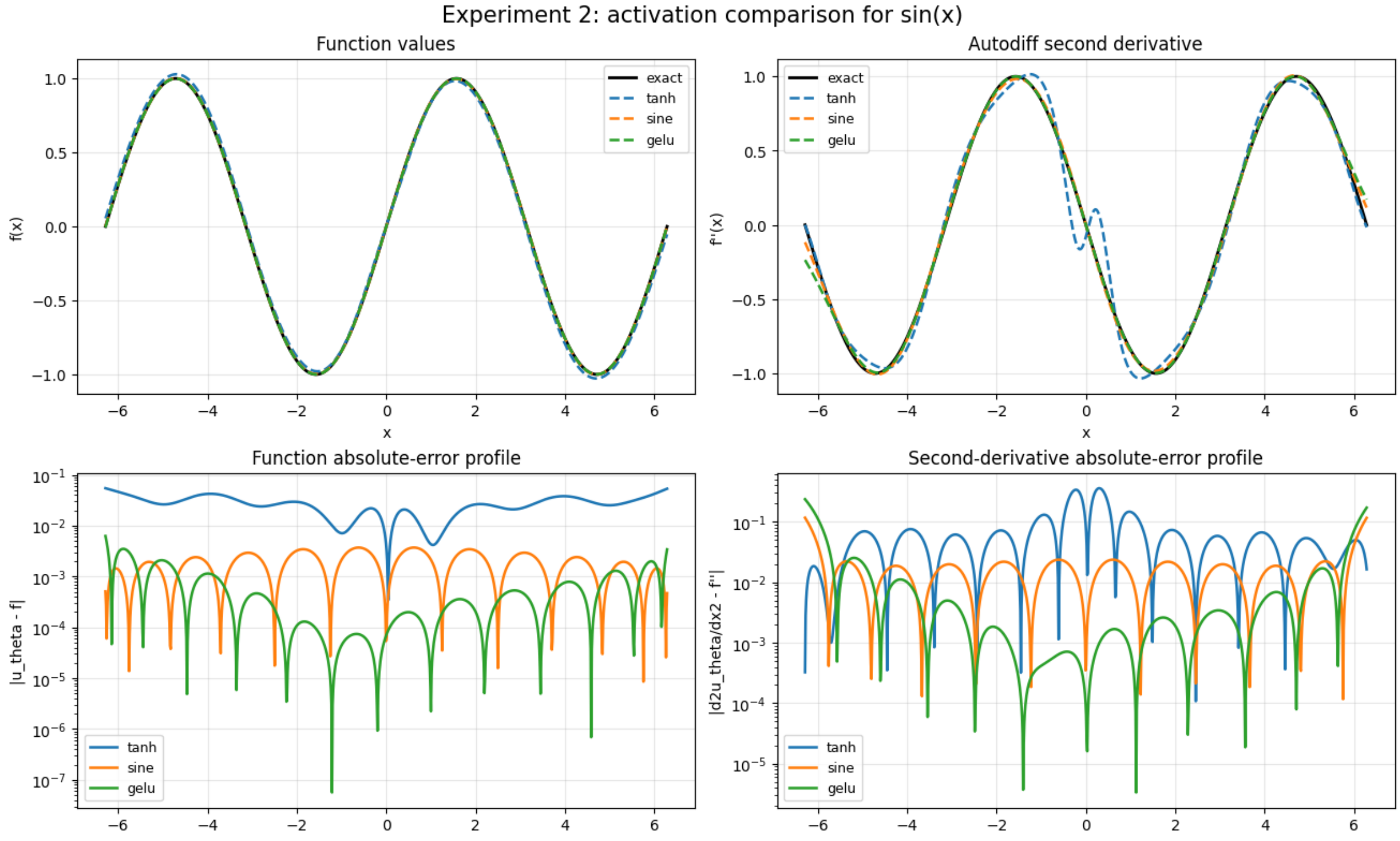

Fig. 3: Activation comparison for sin(x), showing that derivative behavior can differ even when value curves look similar.

Figure 3 shows that, for sin(x), tanh produces a visible local distortion in the second derivative near the center of the domain, whereas sine and gelu produce smoother derivative profiles. The metric table confirms that the activation that is most favorable for function values is not necessarily the same as the activation that is most favorable for second-derivative fidelity.

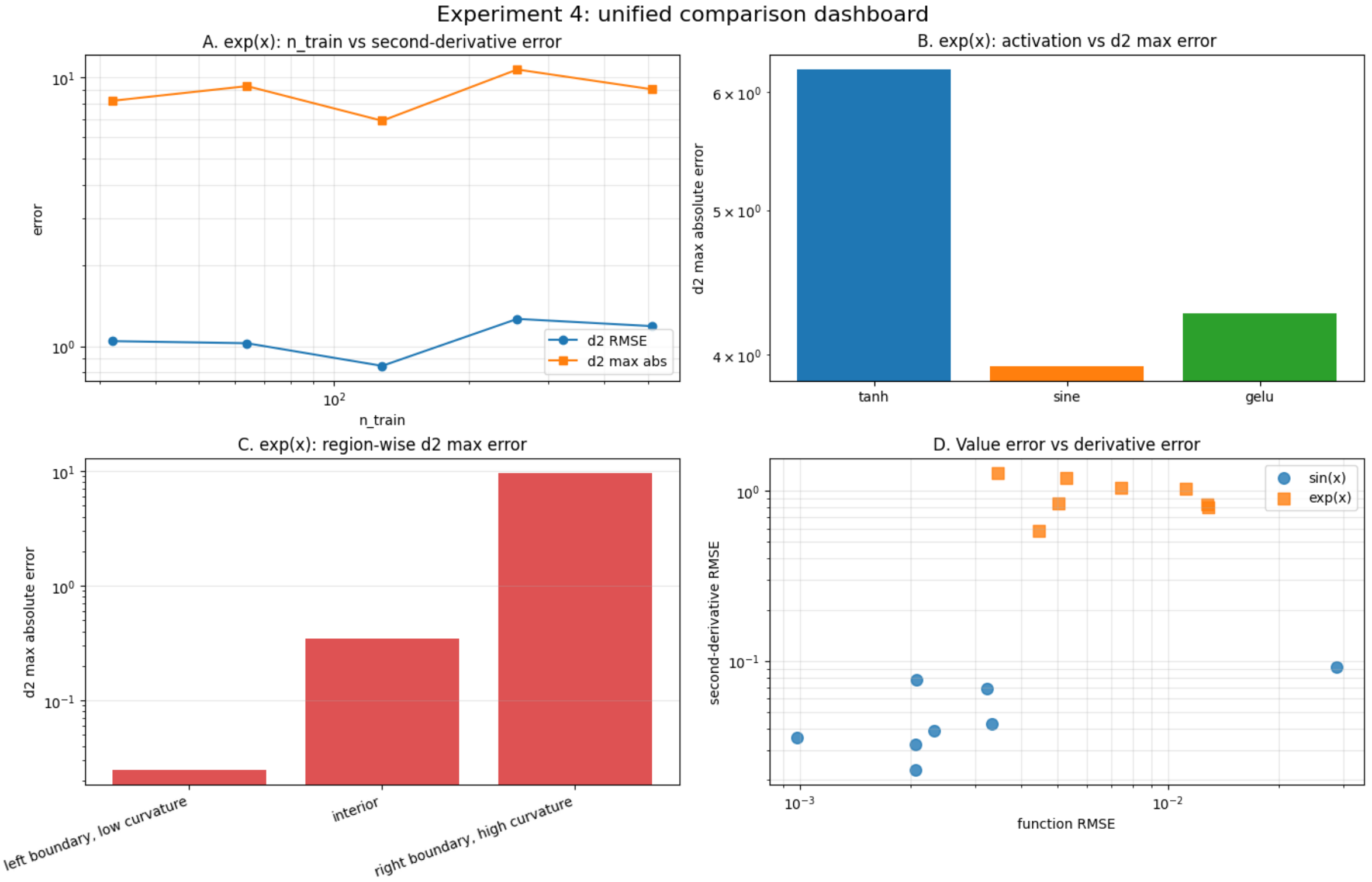

Fig. 4: Unified comparison dashboard for strengthened validation experiments.

Figure 4 summarizes the central evidence. Panel A shows that increasing n_train does not reliably eliminate the exp(x) second-derivative error. Panel B shows that activation selection improves but does not remove the maximum derivative error. Panel C localizes the dominant maximum error to the right high-curvature boundary. Panel D shows a separation between value error and derivative error: exp(x) points retain low function RMSE but high second-derivative RMSE, while sin(x) points occupy a much lower derivative-error range.

## 5. Discussion

The experiment supports the hypothesis that value-level approximation should not be used as a surrogate for derivative-level reliability. This does not contradict derivative approximation theorems for suitable network classes and smooth targets [6]. Rather, it emphasizes that finite training with a value-only objective, finite sampling, a fixed architecture, and practical optimization can leave derivative errors that are large enough to matter for simulation.

The strengthened validation narrows the explanation of the observed failure. The $n_{train}$ sweep shows that the issue is not merely a lack of training points. The activation comparison shows that derivative fidelity is architecture- and activation-sensitive. The endpoint-dense exp(x) evaluation shows that the largest error is localized where boundary effects and high curvature coincide. The maximum-error metric shows that localized failures can remain severe even when RMSE or visual function plots appear acceptable.

For PINNs, the implication is direct. A PDE residual containing $u_{xx}$ can be inaccurate even when u itself appears accurate. This can produce a misleading diagnostic situation: data loss and boundary loss may decrease, plotted solution fields may look plausible, but the residual field can remain wrong in regions where high-order derivatives are poorly represented.

The exp(x) case suggests that high-curvature and boundary regions deserve special attention. In PDE settings, analogous regions may occur near boundary layers, shocks after smoothing, stiff reaction zones, or regions of sharp gradients. The proposed diagnostic is therefore to monitor derivative errors, when analytical derivatives are available, or to monitor residual-error profiles and local operator consistency when they are not.

The present findings also provide a possible interpretation of previously reported structure-preservation failures in standard PINNs. If derivative fidelity deteriorates despite accurate function approximation, residual fields obtained through automatic differentiation may become unreliable, which may in turn contribute to distorted physical structures observed in practical PDE simulations. In this sense, derivative fidelity may represent one possible underlying mechanism underlying structure-preservation problems.

## 6. Limitations and future work

The present study remains deliberately minimal. It uses scalar one-dimensional functions and a small MLP family, and it does not constitute a general impossibility theorem. Adaptive activations and differential-equation software frameworks provide relevant directions for broader testing [7, 8]. The present results should be interpreted as diagnostic evidence that finite value-only training can leave large derivative errors, not as a claim that derivative fidelity cannot be achieved under stronger training objectives or architectures.

A natural next step is to repeat the experiment on simple PDE benchmarks, such as Poisson or heat equations with known analytical solutions, and to compare solution error against residual error. Additional follow-up experiments should include derivative-augmented losses, Sobolev-type training, collocation resampling near high-curvature regions, and explicit comparisons of residual RMSE and residual maximum error.

## 7. Conclusions

This paper identified derivative fidelity as a PINN failure mode and examined it using sin(x) and exp(x) benchmarks. The original value-only experiments showed that accurate-looking function plots can coexist with substantially larger second-derivative errors. The strengthened validation further showed that this gap is not eliminated by simply increasing training points, is sensitive to activation choice, and is localized most severely at the high-curvature right boundary of exp(x).

Therefore, PINN validation should separate value accuracy from derivative and residual accuracy. For simulation reliability, derivative-error profiles, endpoint or high-curvature diagnostics, and maximum absolute residual/error metrics should be treated as first-class diagnostics rather than secondary checks.

## Acknowledgement

During the preparation of this manuscript draft, the author(s) used OpenAI ChatGPT to assist with drafting, organization, and document generation. After using this tool, the author(s) should review and edit the content as needed and take full responsibility for the content of the publication. [Add funding and institutional acknowledgements here, if applicable.]